\documentclass[letterpaper]{article} 
\usepackage[]{aaai23}  
\usepackage{times}  
\usepackage{helvet}  
\usepackage{courier}  
\usepackage[hyphens]{url}  
\usepackage{graphicx} 
\usepackage{natbib}  
\usepackage{caption} 
\usepackage{algorithm}
\usepackage{algorithmic}
\usepackage{amsmath}
\usepackage{amssymb}
\usepackage{bm}
\usepackage{multirow}
\usepackage[dvipsnames,table]{xcolor}

\usepackage{newfloat}
\usepackage{listings}
\DeclareCaptionStyle{ruled}{labelfont=normalfont,labelsep=colon,strut=off} 
\floatstyle{ruled}
\newfloat{listing}{tb}{lst}{}
\floatname{listing}{Listing}
\title{Multi-Region Transfer Learning for Segmentation of Crop Field Boundaries in Satellite Images with Limited Labels}
\author{
    Hannah Kerner\textsuperscript{\rm 1}, Saketh Sundar\textsuperscript{\rm 2}, Manthan Satish\textsuperscript{\rm 1}
}
\affiliations{
    \textsuperscript{\rm 1}School of Computing and Augmented Intelligence, Arizona State University, Tempe, AZ\\
    \textsuperscript{\rm 2}River Hill High School, Clarksville, MD, 21029

}

\usepackage{bibentry}

\begin{document}

\maketitle

\begin{abstract}
The goal of field boundary delineation is to predict the polygonal boundaries and interiors of individual crop fields in overhead remotely sensed images (e.g., from satellites or drones). Automatic delineation of field boundaries is a necessary task for many real-world use cases in agriculture, such as estimating cultivated area in a region or predicting end-of-season yield in a field. Field boundary delineation can be framed as an instance segmentation problem, but presents unique research challenges compared to traditional computer vision datasets used for instance segmentation. The practical applicability of previous work is also limited by the assumption that a sufficiently-large labeled dataset is available where field boundary delineation models will be applied, which is not the reality for most regions (especially under-resourced regions such as Sub-Saharan Africa). We present an approach for segmentation of crop field boundaries in satellite images in regions lacking labeled data that uses multi-region transfer learning to adapt model weights for the target region. We show that our approach outperforms existing methods and that multi-region transfer learning substantially boosts performance for multiple model architectures. Our implementation and datasets are publicly available to enable use of the approach by end-users and serve as a benchmark for future work.
\end{abstract}

\section{Introduction}
\label{sec:intro}

The goal of field boundary delineation is to predict the polygonal boundaries and constituent areas of individual crop fields in overhead remotely sensed images (e.g., from satellites or drones). This can be categorized as a sub-problem within the more general task of instance segmentation in computer vision in which the goal is to detect and delineate the extent of individual occurrences of objects of interest in an image.
Field boundary delineation is an interesting task for studying segmentation in an applied scenario. It is a relevant task for many real-world use cases including estimation of cultivated area, guiding sampling strategies for ground-based surveys \cite{bush1993area,cotter2010area}, and field-scale estimates of quantities such as crop yield, sowing date, or nutrient deficiency \cite{sadeh2019sowing}. Thus there is high potential for adoption of effective solutions by end-users and real world impact in global agriculture and food security \cite{nakalembe2022africa}. 

\begin{figure}[t]
\centering
\includegraphics[width=\columnwidth]{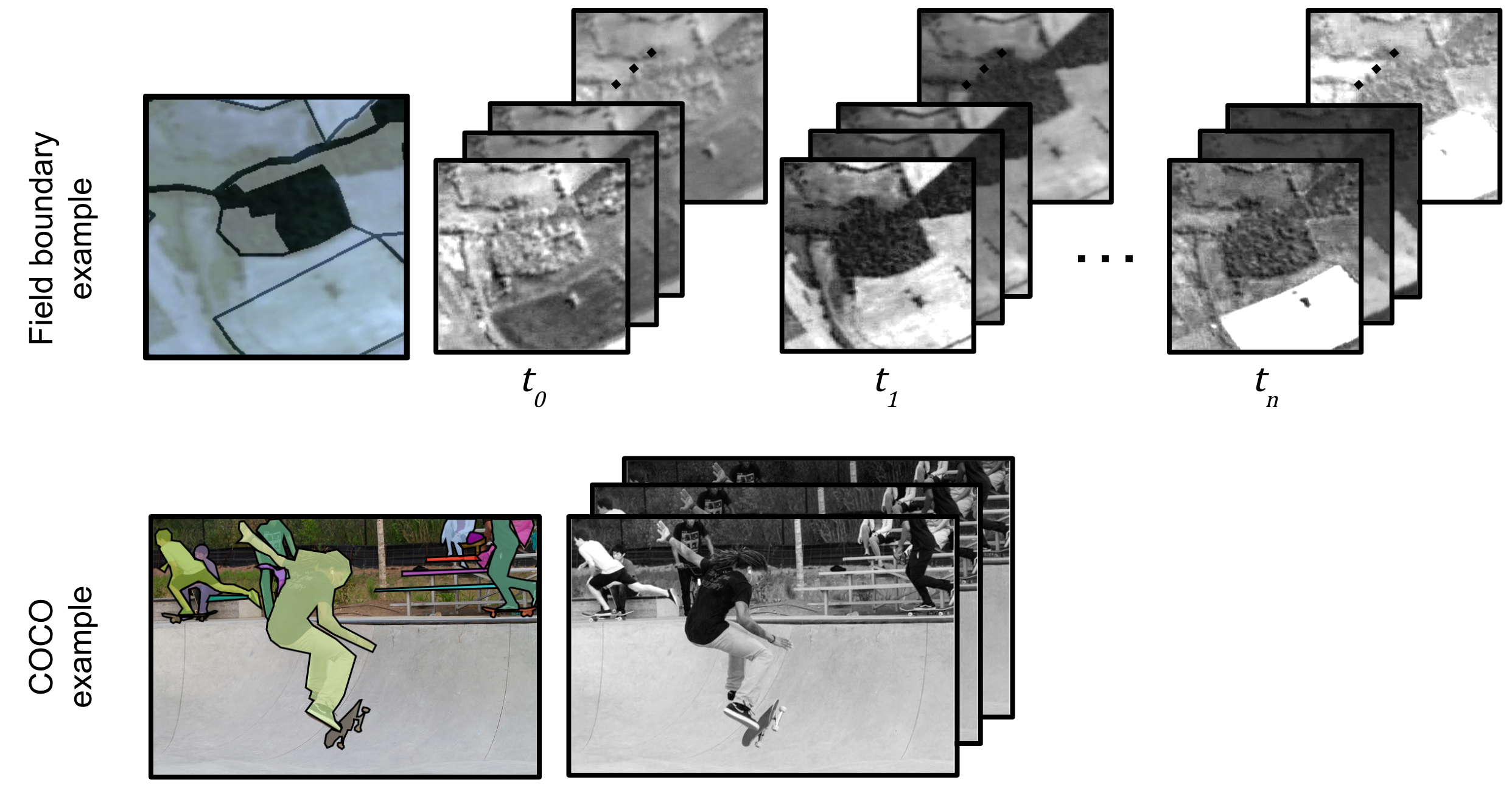} 
\caption{Comparison between instance segmentation of crop fields in satellite remote sensing images (top) and example image from COCO dataset (bottom). In the top row, $t_0$, $t_1$, and $t_n$ represent satellite images acquired at different dates during the year. Top image shows field boundary instances in transparent white and bottom image shows object instance (person) in transparent yellow.}
\label{fig:comparison}
\end{figure}

In addition, field boundary delineation presents several unique challenges and opportunities that are not present in the instance segmentation tasks and datasets commonly studied in computer vision (Figure \ref{fig:comparison}). These challenges can stimulate novel research investigations on foundational and applied topics. One important difference is that the temporal dimension is important for identifying field instances, but most prior work on segmentation does not consider time series of images. Field boundaries are sometimes only clear when inspecting how an image changes over time (e.g., the cycle of a growing season), thus the temporal dimension is an important component of the input. Satellite time series are also different from video datasets since objects are positionally static in satellite images \cite{garnot2021panoptic}. While image datasets include only the red, green, and blue color channels, remote sensing image datasets typically provide four (red, green, blue, and near-infrared) or more spectral channels that capture near-infrared, shortwave infrared, thermal, radar, or other wavelengths in the electromagnetic spectrum. These channels can be equally or more important as the visible channels for segmentation. In remote sensing image datasets, labels are often noisy, sparsely distributed geographically, and images only partially labeled. 

Another important difference is that field boundary delineation typically aims to detect just one class of interest: crop fields. The number and density of field instances present in each image is high compared to traditional image datasets and the size of instances in the image is small (e.g., see Figure \ref{fig:comparison}). There is high variance within this single class in the spatial, spectral, and temporal dimensions due to the variety of crops types grown, field shapes, field sizes, farming practices, and climate and weather patterns observed globally. At the same time, there is low inter-class variance between crop fields and other objects in the non-crop regions of remote sensing images such as forest stands or land parcels. Distribution shift is an omnipresent challenge due to the substantial differences in how crop fields may appear between different regions, climates, and seasons. 

There are also some unique opportunities presented by remote sensing datasets. Unlabeled data that is known to include the class of interest are relatively easy to acquire compared to traditional computer vision datasets. Since the geographic regions supporting crop cultivation are generally known, datasets of images containing crop fields but lacking instance labels are straightforward to construct for a specific region and time period of interest. This presents an opportunity for developing new methods using semi-supervised, self-supervised, or unsupervised approaches for segmentation using remote sensing images.

Most recent studies on field boundary delineation have proposed modified architectures for semantic (rather than instance) segmentation such as U-Nets, which are followed by post-processing steps to isolate individual field instances \cite{wang2022unlocking,aung2020farm,waldner2020deep,persello2019delineation}. Instance segmentation methods like Mask R-CNN have also been adapted \cite{meyer2020deep}. However, the practical applicability of most previous work is limited by the assumption that a sufficiently-large labeled dataset is available where field boundary delineation models will be implemented, which is not the reality for most regions (especially under-resourced regions such as Sub-Saharan Africa) \cite{nakalembe2022africa}. In this study, we present an approach for segmentation of crop field boundaries in satellite images in regions lacking labeled data that uses multi-region transfer learning to adapt model weights for the target region. We additionally provide open datasets and code to stimulate future work on research challenges motivated by the task of field boundary delineation which have been under-studied thus far in segmentation research.

\section{Related work}

Field boundary delineation can be framed as an instance segmentation task in which the goal is to detect and delineate each instance of a crop field that is present in one or more satellite images. Many solutions proposed for this task use traditional unsupervised segmentation algorithms that involve detection of edges or contours and followed by grouping operations (e.g., \citet{yan2014automated,north2018boundary,thomas2020fusion,estes2021high}). For example, \citet{yan2014automated} used the variational region-based geometric active contour (VRGAC) algorithm \cite{chan2001active} to detect candidate field instances in a crop probability map that were further refined using a series of grouping and filtering operations. One limitation of this method is that it requires a crop probability map, which can be difficult to obtain particularly in label-limited regions like Sub-Saharan Africa. \citet{estes2021high} used a multi-step segmentation algorithm to segment candidate field instances in a satellite image, merge the candidate fields with the binary crop classification result from the same satellite image, and refine the polygons by removing holes and smoothing boundaries. While this class of methods has advantages of being transparent, interpretable, and mostly unsupervised,  limitations include inconsistent performance across the diverse appearances of crop fields found globally and that some require additional data sources (e.g., a crop probability or classification map) that do not exist and can be difficult to obtain for many regions.

More recent studies have proposed deep learning solutions with the goal of learning more robust spatial and temporal features for segmenting field instances compared to traditional approaches. While many solutions have been proposed for semantic, instance, and panoptic segmentation in the computer vision literature, there are several key differences between traditional computer vision datasets used in these studies and the agricultural satellite datasets used for field boundary segmentation, as discussed in Introduction and in \citet{garnot2021panoptic}. Prior studies have sought to fill these gaps with specialized architectures or augmentations to existing methods that improve performance on field boundary segmentation in satellite images. \citet{persello2019delineation} used a fully-convolutional network based on SegNet \cite{badrinarayanan2017segnet} to detect field contours which were refined in a series of grouping operations. \citet{aung2020farm} proposed a spatio-temporal U-net to learn temporal patterns useful for detecting individual field instances from satellite images acquired at multiple times of the year; we used this architecture as the starting point for our proposed method. \citet{garnot2021panoptic} proposed a panoptic segmentation approach that introduced convolutional temporal attention for segmenting field instances in a long sequence of satellite images (38-61 timesteps), which enabled the model to learn richer temporal features than could be learned from only one or a few timesteps. \citet{garnot2021panoptic} also detected instances directly rather than performing semantic segmentation followed by additional steps to extract individual instances. \citet{meyer2020deep} also detected instances directly but unlike \citet{garnot2021panoptic} did not use multi-temporal images; they modified Mask R-CNN to detect a larger number and larger range of sizes of candidate objects in the region proposal stage. \citet{waldner2020deep} proposed a multi-task learning solution that improved semantic segmentation performance by simultaneously predicting three related outputs---the extent of fields, field boundaries, and distance to the closest boundary. 

Most prior approaches assume that a sufficiently-large labeled dataset is available for training a model for field boundary segmentation in a region of interest, but this is rarely a reality as discussed in the introduction. To address the limitation of limited labeled data in some regions, \citet{wang2022unlocking} proposed to use transfer learning to pre-train a model on a large dataset from one region and fine-tune it using partial labels from a label-limited region with weak supervision. \citet{wang2022unlocking} is the most similar to our study, but there key differences in our proposed solution. Our solution employs multi-region transfer learning wherein no labels are available for the target (test) region, makes use of coarser-resolution images that are freely available but in which field boundaries are more difficult to detect, and does not assume that a binary crop mask is available for masking out non-crop pixels prior to segmentation.

\section{Methods}
\label{sec:methods}

\begin{figure}[t]
\centering
\includegraphics[width=\columnwidth]{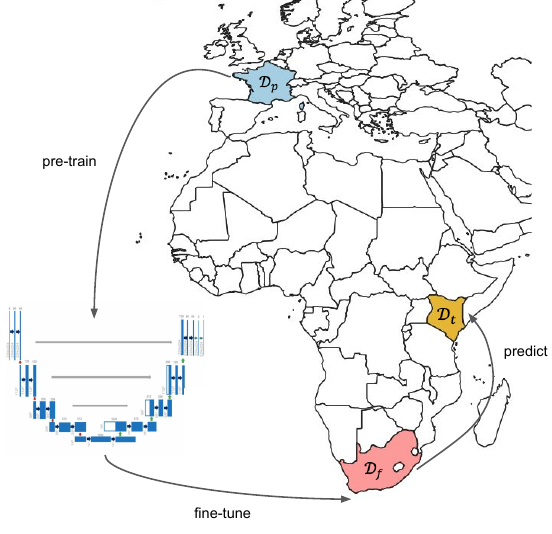} 
\caption{Illustration of multi-region transfer learning approach. Model architecture adapted from \cite{aung2020farm}.}
\label{fig:architecture}
\end{figure}

\subsection{Multi-region transfer learning}
In most transfer learning approaches, a model is pre-trained using a large dataset that may or may not be directly related to the target task (such as ImageNet \cite{yosinski2014transferable}). The model is then fine-tuned by freezing the shallow layers and further training the deeper layers of the network using a smaller labeled dataset from the target task which is drawn from the same distribution as the target or test dataset. We will refer to these datasets as $\mathcal{D}_p$ (the pre-training dataset), $\mathcal{D}_f$ (the fine-tuning dataset), and $\mathcal{D}_t$ (the target or test dataset). A common challenge in real-world scenarios is that there are not sufficient labeled samples available for constructing both a fine-tuning dataset and a test dataset, since few or no labels may be available for the target task. 
For field boundary delineation and other tasks using remote sensing or geospatial data, there are commonly geographic and socioeconomic stratifications in label availability---for example, large labeled datasets tend to be available for high-income countries, medium-sized datasets in medium income countries, and small or no labeled datasets available in low-income countries. This means that some countries are not able to benefit from machine learning and satellite technologies that could help with agricultural monitoring and mitigating food security, among other things. We aim to propose a solution for this label-limited scenario using an approach we term multi-region transfer learning (Figure \ref{fig:architecture}). In multi-region transfer learning, we pre-train a model using a large labeled dataset available from one region ($\mathcal{D}_p$), then fine-tune the model using a smaller dataset available from another region ($\mathcal{D}_f$). Finally, the model is evaluated using a small dataset of labels available for the target region ($\mathcal{D}_t$). The fine-tuning dataset $\mathcal{D}_f$ should share some similarities in object appearance with both the pre-training and target datasets (e.g., similar agricultural practices or growing seasons), and thus can be thought of as a ``bridge'' between the very different pre-training and target datasets. 

\subsection{Model architecture}
Multi-region transfer learning can be implemented using any model architecture. We used the Spatio-Temporal U-net (ST-U-net) architecture proposed by \citet{aung2020farm} as the starting point for this study. This consists of an encoder-decoder architecture with an additional $1\times1$ convolution layer following the input layer to reduce the dimension of the multi-timestep and/or multi-spectral input ($\bm{X} \in \mathbb{R}^{N\times N\times M\times T}$ where $N\times N$ is the image size in pixels, $M$ is the number bands, and $T$ is the number of timesteps) to three bands to match the input dimension of common backbone architectures ($\bm{X} \in \mathbb{R}^{N\times N\times 3} $). We used a ResNet-50 backbone while \citet{aung2020farm} used a ResNet-34. The model is trained to predict two outputs: a field border mask and interior mask. In the border mask, the pixels on the borders of field instances make up the positive class while in the interior mask, the pixels inside the boundaries of the field instances make up the positive class. In both masks, non-field pixels are in the negative class.

\section{Experiments}
\label{sec:experiments}

\subsection{Datasets}
\label{sec:datasets}
We constructed field boundary datasets for three different geographic regions corresponding to the pre-training ($\mathcal{D}_p$), fine-tuning ($\mathcal{D}_f$), and target ($\mathcal{D}_t$) datasets described in Methods. Field boundaries from France, South Africa, and Kenya constitute $\mathcal{D}_p$, $\mathcal{D}_f$, and $\mathcal{D}_t$ respectively. We chose these regions based on the size and quality of publicly-available field boundary datasets as well as their geographic and socio-economic distribution. 

\subsubsection{Instance labels}
We constructed the France dataset from the Registre Parcellaire Graphique, which provides geo-referenced field boundaries for all of France and its territories on an annual basis under an Open License \cite{francefields}. This dataset contains nearly 10 million field instances in each year and is available from years 2010 to present. 
We used only the 2019 dataset and sub-sampled the full dataset to include only the field instances in a dense agricultural region with an area of 17,557 km$^2$ in western France (bounding box coordinates in EPSG:4326 are: $x_{min}=-0.8900$, $x_{max}=0.7673$, $y_{min}=46.0972$, $y_{max}=47.3300$). 

For the South Africa dataset, we used the geo-referenced field boundary labels available from Radiant Earth MLHub \cite{sa-dataset}. This dataset contains field boundaries in the Western Cape region provided by the Western Cape Ministry of Agriculture under a CC BY-NC-SA 4.0 license. The labels were created by manually annotating polygons on aerial images acquired in 2016. A total of 4,151 field labels are provided in the training set. The labels span an area of $543$ km$^2$ (bounding box coordinates in EPSG:4326 are $x_{min}=20.5231$, $x_{max}=20.7896$, $y_{min}=-34.2004$, $y_{max}=-34.0009$).

For the Kenya dataset, we used a dataset of geo-referenced field boundary labels provided by PlantVillage available on Radiant Earth MLHub under a CC-BY-SA-4.0 license \cite{pv-dataset}. These labels were collected during a field survey in 2019; data collectors manually annotated the field boundary labels while they were physically visiting each field using a mobile app in which they could also view their current location on a satellite basemap. The dataset contains 319 total field instances. The labels span an area of $242$ km$^2$ (bounding box coordinates in EPSG:4326 are $x_{min}=34.1644$, $x_{max}=34.3209$, $y_{min}=0.4676$, $y_{max}=0.5933$). Unlike the France and South Africa datasets, in which all field instances are labeled in the region covered by the dataset, the field labels in the Kenya dataset are sparsely distributed across the region. We used the Kenya dataset as a test dataset only. 

\subsubsection{Satellite data}
The field boundary labels described in the previous section are polygons defined by geographic coordinates (latitude, longitude). To create label and input data pairs for machine learning models, one must create a dataset of satellite images covering those geographic coordinates. These image examples can be extracted from a variety of available satellite data sources. We chose to use the Sentinel-2 and PlanetScope image datasets provided by the European Space Agency (ESA) and private company Planet, Inc. respectively. Sentinel-2 data are freely available to the public while PlanetScope data must be obtained through a paid license. Sentinel-2 acquires images with ground resolution of 10 to 60 meters per pixel (i.e., each pixel represents an area between $10\times10$ m$^2$ and $60\times60$ m$^2$ on the ground) with a 5-day revisit time (i.e., each location is nominally imaged every 5 days at the equator and more frequently at the poles). Sentinel-2 provides multispectral images with 13 bands spanning visible, near-infrared, and shortwave infrared wavelengths; in this study we only used the visible bands since these have the highest resolution of 10 m/pixel. PlanetScope images have approximately 3.7 m/pixel ground resolution and a daily revisit time. We used the PSScene4Band Analytic Surface Reflectance product which provides 4-band multispectral images with red, green, blue, and near-infrared channels. We used the 3 visible bands to construct the input images.  

\begin{figure}[t]
\centering
\includegraphics[width=\columnwidth]{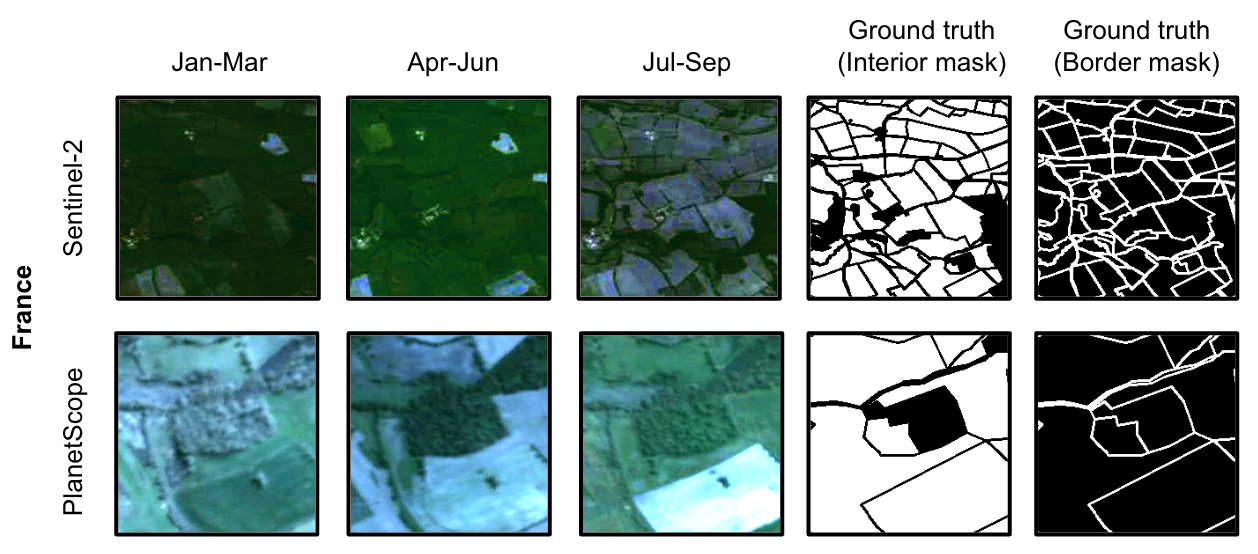} 
\caption{Example images and ground-truth masks from Sentinel-2 and PlanetScope images from France dataset.}
\label{fig:datasets}
\end{figure}

We used the Google Earth Engine python API \cite{gorelick2017google} to pre-process both satellite datasets into image examples with a shape and format amenable to machine learning pipelines. We used the Planet Orders API to deliver the relevant images to Google Earth Engine; the entire Sentinel-2 data archive is already hosted on Google Earth Engine. For each dataset, we created three ``seasonal composite'' images: the seasonal composite contains the median value in each pixel across all images available for a given date range. We used the date ranges January 1 to March 31, April 1 to June 30, and July 1 to September 30 (the year was chosen to be the same year as the label dataset year). It is important when creating satellite datasets to use satellite images with acquisition dates consistent with the label acqusition date because land cover and land use are not static in time---the exact boundary of a field may be different from year to year, or a location may contain a field in one year but another class (e.g., forest) in another year. For example, the France field labels are valid for 2019, so we used satellite images acquired during 2019. This pre-processing resulted in three seasonal composites that were generally cloud-free and showed broad changes in vegetation between each season (see Figure \ref{fig:datasets}). Finally, we tiled the large-area composites into $224\times224$-pixel images and exported them from Google Earth Engine as TFRecords. 

We partitioned the Sentinel-2 and PlanetScope datasets for each region into training, validation, and test sets using a random sample of 80\%, 10\%, and 10\% of the total images respectively. Table \ref{tab:dataset-size} summarizes the number of images in each partition and the average number of field instances per image for the France, South Africa, and Kenya datasets. These datasets are publicly accessible at (Zenodo link redacted for double-blind review).

\begin{table*}
\begin{center}
\begin{tabular}{|l|cccc|cccc|}
\hline
Dataset & \multicolumn{4}{c|}{Sentinel-2} & \multicolumn{4}{c|}{PlanetScope} \\
 & train & val & test & avg. fields/img & train & val & test & avg. fields/img \\
\hline\hline
France & 4000 & 500 & 500 & 51.87 & 56880 & 7110 & 7110 & 4.56\\
South Africa & 143 & 18 & 18 & 41 & 9196 & 250 & 250 & 2.77 \\
Kenya & -- & -- & 18 & 18.82 & -- & -- & 48 & 6.65 \\
\hline
\end{tabular}
\end{center}
\caption{Number of images and average number of field instances per image in the Sentinel-2 and PlanetScope versions of each dataset.}
\label{tab:dataset-size}
\end{table*}

\subsection{Evaluation metrics}
\label{sec:eval-metrics}
We selected four metrics for evaluating model performance on the test set for each region. These metrics were chosen to evaluate segmentation performance as well as the usefulness of the results to the down-stream user, based on how field boundary segmentation results would be used in a practical application. We describe these metrics below.

\paragraph{Pixel-wise F1 score and overall accuracy} The pixel-wise F1 score and overall accuracy provide a measure of how well the exact locations of the predicted field boundaries and interiors match the ground-truth masks. We computed these metrics for both the boundary and interior mask predictions. 

\paragraph{Mean intersection over union} Intersection over union (IoU) is a commonly used metric for evaluating segmentation performance. The IoU averaged over all images in the test set describes how well the shapes predicted in the predicted segmentation masks overlap with the ground-truth masks. We refer to this as the mean IoU (we note this is different from how mIoU is often defined as the mean IoU score across all classes), which we compute as follows:
\begin{equation}
    mIoU = \frac{1}{N}\sum_i^N\frac{TP_i}{TP_i + FP_i + FN_i}
\end{equation}
where $N$ is the number of examples in the test set, TP$_i$ is the number of true positive pixels in test example $i$, FP$_i$ is the number of false positive pixels, and FN$_i$ is the number of false negative pixels. We computed mIoU for both the boundary and interior masks by comparing the predicted masks with the ground-truth mask pixel-wise.

\paragraph{Precision at 0.95 IoU} In downstream use cases of field boundary delineation models, a very close match of the predicted field instance to the ground-truth field instance is very important. A typical end-user may want to use satellite measurements available within the predicted field boundary to estimate the planting date, harvesting date, expected yield, nutrient deficiency, or another characteristic of the crop growing in the field (e.g., \citet{sadeh2019sowing}). Another common use case is to quantify the distribution of the number of fields and field sizes present in a region, perhaps over time to track consolidation or fragmentation of farming operations \cite{estes2021high}. For these use cases, only very accurate matches are acceptable. This can be expressed in terms of IoU. An IoU score over 0.5 may be considered good for generic objects in typical computer vision datasets, but to end-users of field boundary delineation models, an IoU of 0.7 is just as bad as 0.4. Thus, average precision (AP), which averages the precision at a range of IoU thresholds, is not an ideal metric for comparing the utility of multiple field boundary delineation models for end-users. Precision at only a very high IoU threshold, for example 0.95, would provide a more useful measure of model utility for end-users. For this reason, we chose to report the precision at 0.95 threshold on IoU ($P_{IoU\geq0.95}$) for the experiments in this study. To compute precision of detection of individual field instances, each field instance in the masks need to be compared to the ground-truth instances individually. This requires converting the semantic segmentation masks into instance segmentation masks. We used the Rasterio python library to convert the predicted and ground-truth interior masks to a set of polygons, where each polygon corresponds to a closed shape in the mask and has a unique instance id. Using the \texttt{rasterize()} function in Rasterio, we then converted the polygon shapes back to a mask with the pixel value written for each instance corresponding to the instance id. The $P_{IoU\geq0.95}$ was then computed as:
\begin{equation}
    P_{IoU\geq0.95} = \frac{TP_{0.95}}{TP_{0.95} + FP_{0.95}}
\end{equation}

In the Kenya dataset, image examples are only partially labeled. For each image, we have a no-label mask which indicates the pixels for which a label is not available. The no-label mask has the same dimension as the label mask but its values are 1 if there is a valid label and 0 if there is no label for each pixel. For the boundary/border mask, this means that the positive labels are the field borders and the negative labels are the field interiors; all other pixels have no label and are masked out in the metrics computation using the no-label mask. For the interior mask, the positive labels are the field interiors, the negative labels are the field borders, and all other pixels have no label and are masked out.

\subsection{Experimental protocol}
We conducted experiments to evaluate model performance on each of the France, South Africa, and Kenya test sets under multiple transfer learning scenarios. For each transfer learning experiment, we evaluated the Spatio-temporal U-net (ST-U-net) described in Methods as well as a spatial-only U-net, which uses only the July-September composite image as input (i.e., no temporal dimension). In all experiments, the validation set was used for evaluating performance during training and hyperparameter tuning. We describe the experiments for each dataset below.

\paragraph{France} We conducted two experiments for the France test set. In the first experiment, we trained a model using the France Sentinel-2 training set and evaluated it using the France Sentinel-2 test set. In the second experiment, we trained a model using the France Sentinel-2 training set, fine-tuned using the France PlanetScope dataset, and evaluated using the France PlanetScope test set. The goal of this experiment was to evaluate the effectiveness of model transfer to a different spatial resolution for the same region. For both experiments, we evaluated multiple backbone architectures to evaluate the sensitivity of the results to the choice of backbone. 

\paragraph{South Africa} We conducted three experiments for the South Africa test set. To evaluate our proposed cross-region and cross-sensor approach, we pre-trained a model on the France Sentinel-2 training set, fine-tuned it using the South Africa PlanetScope training set, and evaluated it using the South Africa PlanetScope test set. To evaluate the performance gained by fine-tuning, we also evaluated the South Africa PlanetScope test set performance without the fine-tuning from the previous experiment. Finally, we evaluated the performance gained by pre-training with the France Sentinel-2 training set by conducting an experiment in which the model was trained and evaluated with the South Africa PlanetScope data. We used only the ResNet-50 backbone for these and the Kenya experiments since it had comparable performance to ResNet-101 in the France experiments but required significantly less time to train. 

\paragraph{Kenya} The Kenya dataset contains only a test set, thus we did not conduct any experiments involving training using Kenya data. We conducted three experiments for the Kenya test set. To evaluate the baseline performance without multi-region transfer learning, we trained a model on the France Sentinel-2 training set and evaluated its performance on both the Kenya Sentinel-2 and PlanetScope test sets without any fine-tuning. To evaluate the performance of our proposed multi-region, cross-sensor transfer learning approach, we pre-trained a model on the France Sentinel-2 training set, fine-tuned using the South Africa PlanetScope training set, and evaluated using the Kenya PlanetScope test set. 

\paragraph{Baselines} For the experiments that did not use transfer learning (labeled ``no fine-tune'' in Table \ref{tab:results}), we included the method from \citet{aung2020farm} as a baseline comparison. The models used in \citet{aung2020farm} have the same architecture as ours, except \citet{aung2020farm} used a ResNet-34 backbone. For the France dataset, we also evaluated shallower and deeper ResNet backbones (ResNet-18 and ResNet-101). We also evaluated an unsupervised method from as a non-learning baseline for the test sets in all three regions. We implemented the best-performing method from \citet{watkins2019comparison} which delineates field boundaries using a combination of Canny edge detection and watershed segmentation method followed by a rule-set to discard uncultivated areas and reduce noise. This method produces a segmentation mask for the field boundary (border) and does not produce an interior mask as in the other methods.

\section{Results}
\label{sec:results}

\subsection{Quantitative results}
Table \ref{tab:results} summarizes the results from the experiments described in Experiments. 
Across all experiments in each location, ST-U-net models performed better than a U-net model with the same backbone in all metrics. Inputting images from three different times in the ST-U-net instead of one in the standard U-net significantly improved performance. The performance of models improved as the number of layers in the ResNet backbone increased for all datasets, but deeper models also required significantly more training time than shallower models. Across all experiments, performance was substantially higher for the interior prediction task compared to the border prediction task. The unsupervised Canny + watershed method \cite{watkins2019comparison} had the lowest performance for all experiments. 
The effect of fine-tuning is demonstrated in the South Africa and Kenya experiments. Models pre-trained with the France (Sentinel-2) training set and fine-tuned with the South Africa (PlanetScope) training set performed significantly better across all metrics compared to experiments without fine-tuning. Fine-tuning the model with examples from South Africa improved F1 score by 9\% and 16\% for the border and interior predictions respectively, accuracy by 6\% and 13\%, mIoU by 17\% and 5\%, and $P_{IoU\geq0.95}$ by 18\% and 22\% on the Kenya test set for the ST-U-net with ResNet-50 backbone. When the fine-tuning and test dataset were both from South Africa, fine-tuning improved F1 score by 18\% and 30\% for border and interior predictions respectively, accuracy by 18\% and 26\%, mIoU by 14\% and 19\%, and $P_{IoU\geq0.95}$ by 17\% and 29\% for the ST-U-net with ResNet-50 backbone.

\setlength{\tabcolsep}{1pt}
\begin{table*}
\scriptsize
\begin{center}
\begin{tabular}{|ll|cccc|cccc|}
\hline
& & \multicolumn{4}{c|}{Border} & \multicolumn{4}{c|}{Interior} \\
 & & F1 & acc & mIoU & $P_{IoU\geq0.95}$ & F1 & acc & mIoU & $P_{IoU\geq0.95}$ \\
\hline\hline
\multirow{16}{*}{\rotatebox[origin=c]{90}{France}} & France (S2) $\rightarrow$ France (S2) (no finetune) &  &  &  &  &  &  &  &  \\
& \hspace{2mm} ST-U-net (Resnet-18)  & $0.56 \pm 0.02$ & $0.72 \pm 0.01$ & $0.71 \pm 0.01$ & $0.42 \pm 0.01$ & $0.71 \pm 0.01$ & $0.73 \pm 0.02$ & $0.80 \pm 0.01$ & $0.72 \pm 0.02$ \\ 
& \hspace{2mm} ST-U-net (ResNet-50)  & $0.71 \pm 0.01$ & $0.87 \pm 0.01$ & $0.81 \pm 0.01$ & $0.59 \pm 0.02$ & $\bm{0.88 \pm 0.01}$ & $\bm{0.89 \pm 0.01}$ & $\bm{0.93 \pm 0.01}$ & $0.87 \pm 0.01$ \\ 
& \hspace{2mm} ST-U-net (Resnet-101) & $\bm{0.75 \pm 0.01}$ & $\bm{0.90 \pm 0.02}$ & $\bm{0.83 \pm 0.01}$ & $\bm{0.66 \pm 0.01}$ & $\bm{0.89 \pm 0.01}$ & $\bm{0.91 \pm 0.01}$ & $\bm{0.95 \pm 0.01}$ & $\bm{0.91 \pm 0.02}$ \\  
& \hspace{2mm} ST-U-net (ResNet-34)$^1$ & $0.56 \pm 0.02$ & $0.78 \pm 0.01$ & $0.75 \pm 0.01$ & $0.48 \pm 0.01$ & $0.81 \pm 0.01$ & $0.82 \pm 0.01$ & $0.86 \pm 0.01$ & $0.78 \pm 0.01$ \\
& \hspace{2mm} U-net (Resnet-18) & $0.51 \pm 0.01$ & $0.66 \pm 0.01$ & $0.64 \pm 0.01$ & $0.23 \pm 0.02$ & $0.66 \pm 0.02$ & $0.67 \pm 0.02$ & $0.70 \pm 0.01$ & $0.51 \pm 0.01$ \\ 
& \hspace{2mm} U-net (ResNet-50) & $0.69 \pm 0.01$ & $0.83 \pm 0.02$ & $0.76 \pm 0.01$ & $0.47 \pm 0.01$ & $0.82 \pm 0.01$ & $0.83 \pm 0.01$ & $0.87 \pm 0.01$ & $0.78 \pm 0.01$\\ 
& \hspace{2mm} U-net (Resnet-101) & $0.72 \pm 0.01$ & $\bm{0.88 \pm 0.01}$ & $0.80 \pm 0.01$ & $0.54 \pm 0.01$ & $0.85 \pm 0.02$ & $0.88 \pm 0.01$ & $0.89 \pm 0.01$ & $0.82 \pm 0.01$ \\ 

& France (S2) $\rightarrow$ France (PS) (PS finetune) &  &  &  &  &  &  &  &  \\ 
& \hspace{2mm} ST-U-net (Resnet-18) & $0.50 \pm 0.03$ & $0.63 \pm 0.02$ & $0.59 \pm 0.01$ & $0.16 \pm 0.01$ & $0.67 \pm 0.02$ & $0.67 \pm 0.01$ & $0.68 \pm 0.01$  & $0.35 \pm 0.02$\\ 
& \hspace{2mm} ST-U-net (ResNet-50) & $0.69 \pm 0.01$ & $0.81 \pm 0.01$ & $0.72 \pm 0.01$ & $0.42 \pm 0.01$ & $0.82 \pm 0.01$ & $0.85 \pm 0.02$ & $0.83 \pm 0.01$  & $0.69 \pm 0.02$\\ 
& \hspace{2mm} ST-U-net (Resnet-101) & $\bm{0.73 \pm 0.01}$ & $0.85 \pm 0.01$ & $0.77 \pm 0.01$ & $0.44 \pm 0.01$ & $0.85 \pm 0.01$ & $0.88 \pm 0.01$ & $0.85 \pm 0.01$ & $0.73 \pm 0.01$\\ 
& \hspace{2mm} U-net (Resnet-18) & $0.46 \pm 0.02$ & $0.58 \pm 0.01$ & $0.58 \pm 0.02$ & $0.15 \pm 0.02$ & $0.63 \pm 0.02$ & $0.65 \pm 0.01$ & $0.64 \pm 0.01$  & $0.47 \pm 0.02$\\ 
& \hspace{2mm} U-net (ResNet-50) & $0.66 \pm 0.01$ & $0.78 \pm 0.01$ & $0.69 \pm 0.02$ & $0.37 \pm 0.02$ & $0.74 \pm 0.02$ & $0.77 \pm 0.02$ & $0.79 \pm 0.01$  & $0.68 \pm 0.01$\\ 
& \hspace{2mm} U-net (Resnet-101) & $0.70 \pm 0.01$ & $0.82 \pm 0.02$ & $0.71 \pm 0.01$ & $0.42 \pm 0.01$ & $0.76 \pm 0.01$ & $0.81 \pm 0.01$ & $0.81 \pm 0.01$  & $0.72 \pm 0.02$\\ 

& France S2 (Canny + watershed$^2$) & $0.24$ & $0.04$ & $0.13$ & $0.00$ & --- & --- & --- & --- \\ 

\hline \hline

\multirow{12}{*}{\rotatebox[origin=c]{90}{South Africa (SA)}} & SA (PS) $\rightarrow$ SA (PS) (no finetune) &  &  &  &  &  &  &  &  \\ 
& \hspace{2mm} ST-U-net (ResNet-50) & $\bm{0.74 \pm 0.02}$  & $\bm{0.87 \pm 0.02}$ & $\bm{0.79 \pm 0.01}$ & $\bm{0.57 \pm 0.01}$ & $\bm{0.86 \pm 0.01}$ & $\bm{0.92 \pm 0.01}$ & $\bm{0.91 \pm 0.01}$ & $\bm{0.86 \pm 0.01}$ \\ 
& \hspace{2mm} ST-U-net (ResNet-34)$^1$ & $0.68 \pm 0.02$  & $0.81 \pm 0.01$ & $0.75 \pm 0.01$ & $0.52 \pm 0.02$ & $0.83 \pm 0.02$ & $0.87 \pm 0.02$ & $\bm{0.89 \pm 0.01}$ & $0.83 \pm 0.01$\\ 
& \hspace{2mm} U-net (ResNet-50) & $0.72 \pm 0.01$ & $\bm{0.84 \pm 0.01}$ & $0.75 \pm 0.02$ & $0.51 \pm 0.01$ & $\bm{0.84 \pm 0.01}$ & $0.87 \pm 0.02$ & $\bm{0.89 \pm 0.01}$ & $0.83 \pm 0.02$\\ 

& France (S2) $\rightarrow$ SA (PS) (no finetune) &  &  &  &  &  &  &  &   \\ 
& \hspace{2mm} ST-U-net (ResNet-50) & $0.52 \pm 0.01$ & $0.65 \pm 0.01$ & $0.65 \pm 0.01$ & $0.38 \pm 0.02$ & $0.55 \pm 0.01$ & $0.58 \pm 0.01$ & $0.67 \pm 0.01$ & $0.54 \pm 0.01$ \\ 
& \hspace{2mm} ST-U-net (ResNet-34)$^1$ & $0.44 \pm 0.01$ & $0.54 \pm 0.01$ & $0.56 \pm 0.01$ & $0.32 \pm 0.01$ & $0.48 \pm 0.01$ & $0.51 \pm 0.01$ & $0.62 \pm 0.01$ & $0.50 \pm 0.02$\\
& \hspace{2mm} U-net (ResNet-50) & $0.48 \pm 0.01$ & $0.57 \pm 0.02$ & $0.58 \pm 0.02$ & $0.29 \pm 0.01$ & $0.52 \pm 0.01$ & $0.53 \pm 0.01$ & $0.63 \pm 0.01$ & $0.52 \pm 0.02$\\ 

& France (S2) $\rightarrow$ SA (PS) (SA finetune) &  &  &  &  &  &  &  &  \\ 
& \hspace{2mm} ST-U-net (ResNet-50) & $\bm{0.71 \pm 0.01}$ & $0.82 \pm 0.01$ & $\bm{0.79 \pm 0.01}$ & $\bm{0.57 \pm 0.01}$ & $\bm{0.85 \pm 0.02}$ & $0.84 \pm 0.02$ & $\bm{0.89 \pm 0.01}$ & $\bm{0.84 \pm 0.01}$\\ 
& \hspace{2mm} U-net (ResNet-50) & $0.65 \pm 0.02$ & $0.77 \pm 0.01$  & $0.67 \pm 0.01$ & $0.29 \pm 0.01$ & $0.76 \pm 0.02$ & $0.77 \pm 0.02$ & $0.83 \pm 0.01$ & $0.74 \pm 0.02$ \\ 

& SA PS (Canny + watershed$^2$) & $0.24$ & $0.03$ & $0.12$ & $0.00$ & --- & --- & --- & --- \\ 

\hline \hline

\multirow{12}{*}{\rotatebox[origin=c]{90}{Kenya}} & France (S2) $\rightarrow$ Kenya (S2) (no fine-tune) &  &  &  &  &  &  &  & \\ 
& \hspace{2mm} ST-U-net (ResNet-50) & $0.18 \pm 0.01$ & $0.27 \pm 0.01$ & $0.36 \pm 0.02$ & $0.08 \pm 0.02$ & $0.32 \pm 0.01$ & $0.43 \pm 0.02$ & $0.45 \pm 0.01$ & $0.14 \pm 0.02$ \\ 
& \hspace{2mm} ST-U-net (ResNet-34)$^1$ & $0.13 \pm 0.01$ & $0.21 \pm 0.01$ & $0.31 \pm 0.01$ & $0.04 \pm 0.01$ & $0.26 \pm 0.02$ & $0.35 \pm 0.01$ & $0.41 \pm 0.01$ & $0.11 \pm 0.02$ \\
& \hspace{2mm} U-net (ResNet-50) & $0.18 \pm 0.01$ & $0.24 \pm 0.01$ & $0.32 \pm 0.02$ & $0.02 \pm 0.02$ & $0.28 \pm 0.01$ & $0.38 \pm 0.01$ & $0.41 \pm 0.01$ & $0.11 \pm 0.02$ \\ 

& France (S2) $\rightarrow$ Kenya (PS) (no fine-tune) &  &  &  &  &  &  &  &  \\ 
& \hspace{2mm} ST-U-net (ResNet-50) & $0.29 \pm 0.02$ & $0.39 \pm 0.02$ & $0.42 \pm 0.02$ & $0.11 \pm 0.02$ & $0.37 \pm 0.02$ & $0.48 \pm 0.02$ & $0.59 \pm 0.01$ & $0.29 \pm 0.01$\\ 
& \hspace{2mm} ST-U-net (ResNet-34)$^1$ & $0.26 \pm 0.01$ & $0.37 \pm 0.02$ & $0.35 \pm 0.02$  & $0.07 \pm 0.02$ & $0.34 \pm 0.02$ & $0.43 \pm 0.01$ & $0.54 \pm 0.02$ & $0.25 \pm 0.02$ \\
& \hspace{2mm} U-net (ResNet-50) & $0.27 \pm 0.02$ & $0.36 \pm 0.02$ & $0.37 \pm 0.01$ & $0.08 \pm 0.01$ & $0.34 \pm 0.02$ & $0.45 \pm 0.02$ & $0.57 \pm 0.01$ & $0.26 \pm 0.01$ \\  


\rowcolor{gray!30}& France (S2) $\rightarrow$ Kenya (PS) (SA fine-tune) &  &  &  &  &  &  &  &  \\ 
\rowcolor{gray!30}& \hspace{2mm} ST-U-net (ResNet-50) & $\bm{0.37 \pm 0.01}$ & $\bm{0.45 \pm 0.01}$ & $\bm{0.58 \pm 0.01}$ & $\bm{0.30 \pm 0.01}$ & $\bm{0.53 \pm 0.02}$ & $\bm{0.61 \pm 0.02}$ & $\bm{0.64 \pm 0.01}$ & $\bm{0.51 \pm 0.02}$ \\ 
\rowcolor{gray!30}& \hspace{2mm} U-net (ResNet-50) & $0.33 \pm 0.01$ & $\bm{0.43 \pm 0.02}$ & $0.55 \pm 0.01$ & $0.25 \pm 0.01$ & $\bm{0.52 \pm 0.01}$ & $\bm{0.58 \pm 0.01}$ & $\bm{0.63 \pm 0.01}$ & $0.47 \pm 0.01$ \\ 

& Kenya PS (Canny + watershed$^2$) & $0.25$ & $0.04$ & $0.13$ & $0.00$ & ---- & --- & --- & --- \\ 

\hline
\end{tabular}
\end{center}
\caption{Experiment results for the France, South Africa, and Kenya datasets. Results reported using mean and standard deviation from 10 runs with different random seeds (except for Canny + watershed \cite{watkins2019comparison}, which does not use a random seed). Highest metric for each test region in \textbf{bold}. Our proposed multi-region transfer learning method is shaded in gray for the Kenya dataset. ``SA fine-tune'' indicates that the model was fine-tuned with South Africa training data. \\$^1$ indicates architecture proposed by \cite{aung2020farm}. \\$^2$ indicates algorithm proposed by  \cite{watkins2019comparison}.}
\label{tab:results}
\end{table*}

\begin{figure}[t]
\begin{center}
\includegraphics[width=\columnwidth]{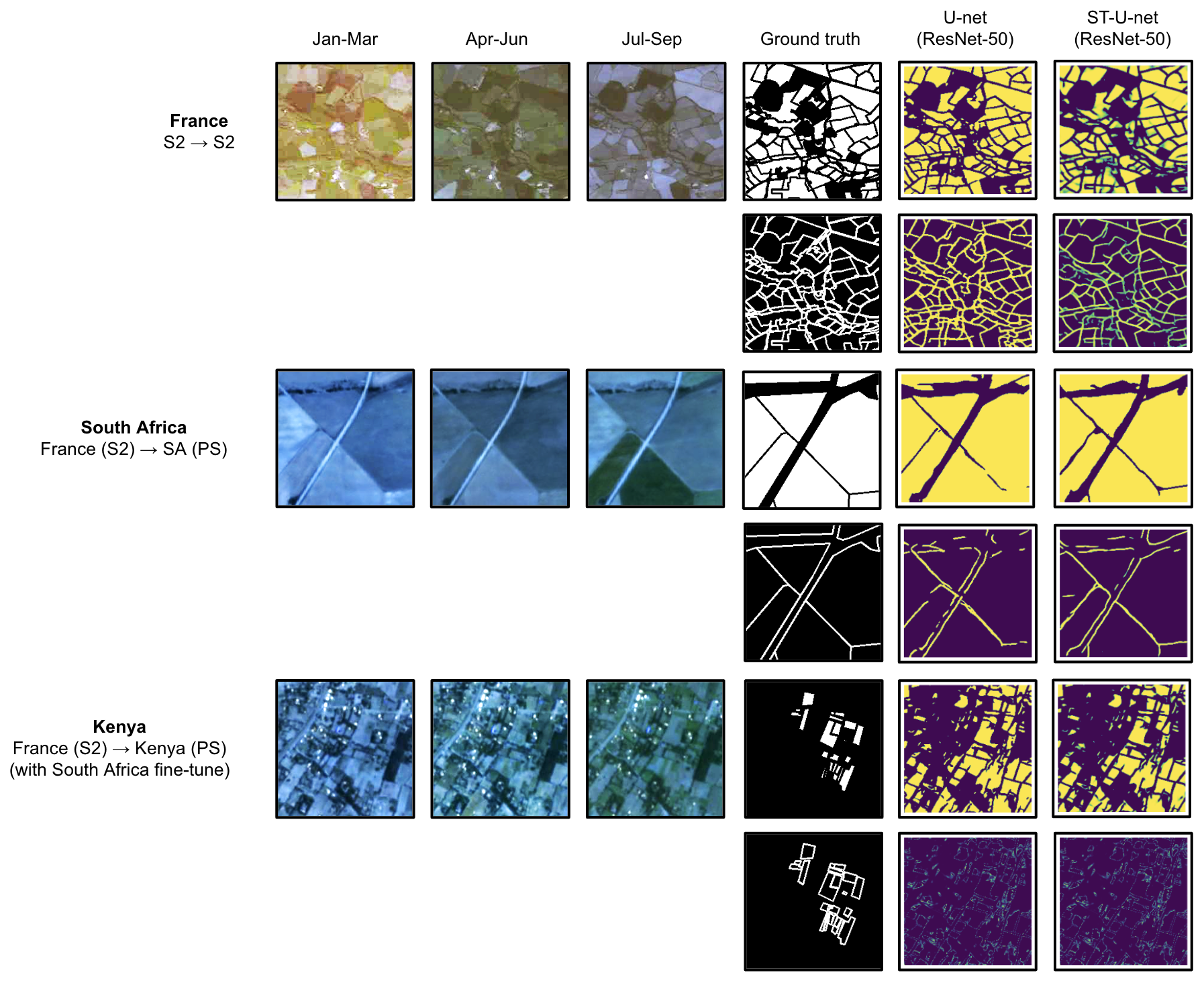}
\end{center}
  \caption{Qualitative field boundary segmentation results for example test images in selected experiments from Table \ref{tab:results}.}
\label{fig:qualitative-results}
\end{figure}

\subsection{Qualitative results}
Figure \ref{fig:qualitative-results} shows example qualitative results from selected experiments for each dataset. Consistent with the quantitative results, these examples show that the ST-U-net results more closely match the ground-truth masks in both the border and interior predictions. The ST-U-net appears to more accurately predict fully-closed field boundaries, as disconnected boundary lines can be seen more frequently in the standard U-net predictions which has the most negative effect on the $P_{IoU\geq0.95}$ metric. The Kenya example illustrates the partially-labeled images in that dataset. Though only some fields in the center of the image are labeled, both models predict many more fields that can also be seen in the input images. The performance of the border prediction task is substantially lower for the Kenya dataset compared to the France and South Africa datasets for both models.

\section{Discussion}
The experimental results summarized in Results show that multi-region transfer learning and multi-temporal image input sequences can significantly improve field boundary segmentation performance, particularly when the goal is to make predictions for a region lacking labeled datasets for training. In this study we incorporated temporal information by stacking images from three time periods in the input and reducing the dimension in the first layer of the ST-U-net. It is possible that these results could be further improved by using other recent models that more explicitly model the temporal patterns in the time series, such as \citet{garnot2021panoptic}. In addition, differences in growing seasons between the regions used for multi-region transfer learning could negatively impact performance of our approach. Future studies may improve performance by considering the different growing seasons of crops in the regions captured in the datasets, for example by defining the input time periods by growth stage instead of specific months as in \citet{kerner2022phenological}.

Even though all labels in the Kenya dataset were used for evaluating test set performance, the dataset is still very small (48 images for the PlanetScope dataset; see Table \ref{tab:dataset-size}). It is not feasible or sustainable for future efforts to address this limited labeled data challenge by creating new labeled datasets alone; since abundant unlabeled data can be obtained from remote sensing data sources, future work could instead investigate methods for evaluating performance of models for segmentation and other tasks using unlabeled data.

Agriculture in Kenya is predominantly smallholder farming where fields are typically smaller than 5 hectares \cite{nakalembe2022africa}, thus the size of field instances is substantially smaller in the Kenya dataset compared France and South Africa. The high 3 m/pixel resolution of commercial PlanetScope images is necessary for resolving these small field sizes for segmentation, while the coarser 10 m/pixel resolution of freely-available Sentinel-2 images is sufficient for France and South Africa. An important benefit  of our multi-region transfer learning approach for end-users is that we make efficient use of commercial data by leveraging free data for pre-training, which is the majority of the total data used, and use paid data only for fine-tuning and/or inference.   



\section{Conclusion}

We introduced an approach for segmentation of crop field boundaries in satellite images in regions lacking labeled data that uses multi-region transfer learning to adapt model weights for the target region. Using three datasets from three countries (France, South Africa, and Kenya), we showed that multi-region transfer learning substantially boosts performance for multiple model architectures. Our implementation is available on Github: \url{https://github.com/kerner-lab/transfer-field-delineation}.



\subsubsection{Acknowledgments.}
This work was supported by the NASA Harvest Consortium (Award
Number 80NSSC17K0625), the USDA-Foreign Agricultural Service, and USAID (Award Number FX22TA10960R004, project title Earth Observations for Field Level Agricultural Resource Mapping (EO-Farm): Pilot in Rwanda in Support of NISR).



\bibliography{field-boundaries-refs}



\end{document}